%% file: main.tex
\documentclass[runningheads]{llncs}

\ifdefined\pdfsuppressptexinfo\pdfsuppressptexinfo=-1\fi

\usepackage[T1]{fontenc}
\usepackage{lmodern}
\usepackage{amsmath}
\usepackage{amssymb}
\usepackage{graphicx}
\usepackage{booktabs}
\usepackage{array}
\usepackage{tikz}
\usetikzlibrary{decorations.pathreplacing}
\usepackage{hyperref}

\input{utils}

\title{Optimal Connectivity of Random Boolean Networks for Reservoir Computing}
\titlerunning{Optimal Connectivity of Random Boolean Networks}

\author{Anonymous}
\authorrunning{Anonymous}
\institute{Anonymous}
\ifreview\else\InputIfFileExists{private.tex}{}{\typeout{no private.tex: Anonymous, no links}}\fi
\begin{document}

\maketitle

\input{sections/abstract}

\input{sections/introduction}

\input{sections/background}

\input{sections/method}

\input{sections/results}

\input{sections/discussion}

\input{sections/conclusion}

\bibliographystyle{splncs04}
\bibliography{furu}

\input{sections/appendix}

\end{document}

%% file: utils.tex
\graphicspath{{figures/}}

\newcommand{\Iset}{\ensuremath{I}}
\newcommand{\Rset}{\ensuremath{R}}
\newcommand{\Vset}{\ensuremath{V}}
\newcommand{\nR}{\ensuremath{|\Rset|}}
\newcommand{\kavg}{\ensuremath{\bar{k}}}
\newcommand{\kbest}{\ensuremath{\bar{k}^{*}}}
\newcommand{\kmax}{\ensuremath{k_{\max}}}
\newcommand{\redund}{\ensuremath{r}}
\newcommand{\nticks}{\ensuremath{\nu}}
\newcommand{\selfloop}{\ensuremath{s}}

\newif\ifreview
\ifnum\pdfstrcmp{\jobname}{review}=0 \reviewtrue\fi

\newcommand{\dashfig}[3]{\ifdefined\dashbase\href{\dashbase#1?#2\&expand=false}{#3}\else#3\fi}

\newcommand{\dashavail}[2]{\ifdefined\dashbase#1\else#2\fi}

\newcommand{\codelink}{\ifdefined\codeurl\expandafter\url\expandafter{\codeurl}\else%
  \url{https://anonymous.4open.science/r/754F/}\ifreview{} (anonymized)\fi\fi}

%% file: private.tex
\newcommand*{\codeurl}{https://github.com/chrisvibe/furu}
\author{Christopher Michael Skeide Vibe\inst{1}
\and Aliaksandr Hubin\inst{2}
\and Stefano Nichele\inst{1}}
\authorrunning{C. M. S. Vibe et al.}
\institute{Machine Learning Research Group, \O stfold University of Applied
Sciences, B R A Veien 4, NO-1783 Halden, Norway\\
\email{christopher.vibe@hiof.no}
\and
Bioinformatics and Applied Statistics, Faculty of Chemistry, Biotechnology and
Food Science, Norwegian University of Life Sciences, Chr.\ M.\ Falsens vei 18,
NO-1433 {\AA}s, Norway}

%% file: sections/abstract.tex
\begin{abstract}
Reservoir computing shifts the focus from training a system's internal dynamics to designing dynamics that expose useful computational capabilities via linear projection. We study random Boolean networks as reservoirs: cellular automata with random recurrent wiring and a random rule at each node. Their main parameter is the connectivity, the mean number of edges into a node. Without input, these networks pass from ordered to chaotic dynamics at a connectivity of $2$, where their dynamics are expected to facilitate computation. How the input is connected is itself a design factor: if the input shares the network's edges, the connectivity sets both the dynamics and how many nodes receive the input, the input fan-out. We set the input fan-out separately and compare $7{,}560$ reservoir designs, all combinations of six design factors, each built as many random reservoirs, on recalling the majority or the parity of a few input bits seen several steps earlier. The optimal connectivity falls from $2.5$ to $1.75$--$2$ as more nodes receive the input, but hardly depends on the task or on the other design factors, except self-loops, which connect a node to itself and raise the optimal connectivity. A measure computed from the reservoir states without a trained readout peaks at a connectivity that also falls with the fan-out and rises with self-loops, but does not replace training a readout. Reservoirs of one design vary widely, so a design should be judged by its typical reservoir.
\end{abstract}

%% file: sections/introduction.tex
\section{Introduction}\label{sec:introduction}

In reservoir computing, the input drives a dynamical system, the reservoir, and only a linear readout of its states is trained \cite{jaegerEchoStateApproach2010,maassRealtimeComputingStable2002}. Thus the design of the reservoir limits what the system can compute. Random Boolean networks (RBNs) generalize cellular automata (CA), with random instead of local wiring and a rule drawn for each node. CA with a rule per cell have served as reservoirs \cite{nicheleReservoirComputingUsing2017}, and Snyder et al.\ call RBNs used as reservoirs Boolean reservoirs \cite{snyderFindingOptimalRandom2012b}. Without input, an RBN with unbiased rules is ordered when its connectivity, the mean in-degree $\kavg$ of its nodes, is below $2$ and chaotic above it \cite{kauffmanMetabolicStabilityEpigenesis1969,derridaRandomNetworksAutomata1986}. Reservoirs are expected to perform best near this transition \cite{bertschingerRealtimeComputationEdge2004,legensteinEdgeChaosPrediction2007}.

A reservoir, however, is driven: at every step the input enters the rules of the nodes it reaches, so the critical value of the undriven network need not be the optimal connectivity $\kbest$ of a reservoir, the $\kavg$ at which it is most accurate. Earlier results differ. Snyder et al.\ draw the input and recurrent edges from one shared pool, so that $\kavg$ also sets how many nodes receive the input, the input fan-out $\redund$ (Sect.~\ref{sec:background}); their trained $500$-node reservoirs are most accurate at the largest $\kavg$ tested, $8$, at the shortest delay, and at $\kavg$ between $3$ and $4$ at long delays \cite{snyderFindingOptimalRandom2012b}. In a second study they wire the input to $\redund$ separate nodes and vary $\redund$ and $\kavg \in \{1,2,3\}$; on their tasks, accuracy falls least at $\kavg{=}2$ as the window and the input stream grow \cite{snyderComputationalCapabilitiesRandom2013}. Echlin et al.\ also set the fan-out separately and find $\kbest{=}2$ among $\{1,2,3\}$, except in large reservoirs with many input nodes, where $\kbest{=}1$ \cite{echlinFlexibilityBooleanNetwork2018}; in threshold networks, Calvet et al.\ find $\kbest{=}4$ \cite{calvetConnectivityDegreeControls2024}. Each study varies a few design factors and fixes the rest, so $\kbest$ has not been measured while the input fan-out and other factors vary together.

We therefore ask how $\kbest$ depends on the design of the reservoir and on the task. We expect $\kbest$ to vary with every design factor, the input fan-out among them, and with how much memory and computation the task demands. As in the second study of Snyder et al.\ and in Echlin et al., we give the input its own edges, to $\redund$ reservoir nodes drawn at random, so that the input fan-out is set apart from $\kavg$. We vary six design factors jointly: the input fan-out, $\kavg$, state updates per input bit, initialization, input injection, and self-loops. The classical RBN has no self-loops; Montagna et al.\ replace one in-edge of a node with a self-loop, preserving its random rule, and find that the undriven network has more cyclic attractors the more nodes have a self-loop \cite{montagnaImpactSelfLoopsBoolean2021}. Our self-loops replace an in-edge in the same way, so at a given $\kavg$ a network with self-loops has fewer edges between distinct nodes. In a driven reservoir, the input variance is also expected to affect the dynamics; we hold it fixed (Sect.~\ref{sec:method_tasks}) and vary the demands of the task instead. Each task recalls a short window of past input bits, either its majority or its parity. Majority and parity are two ends of a scale from memory to computation: the majority can still be guessed when some bits of the window are lost, while the parity falls to chance when a single bit is lost. Memory and nonlinear computation share the bounded capacity of a dynamical system \cite{dambreInformationProcessingCapacity2012}.

Boolean reservoirs are often judged without training a readout \cite{snyderFindingOptimalRandom2012b,snyderComputationalCapabilitiesRandom2013}, by two measures computed from the reservoir states alone \cite{legensteinEdgeChaosPrediction2007}. Kernel quality (KQ) is computed by driving the reservoir with streams that differ and counting the linearly independent states that result, so a high value means that distinct input histories give distinct states; generalization rank (GR) uses streams that are identical at selected bit positions, so a low value means that histories which agree there end in similar states. Snyder et al.\ take the $\kavg$ at which the difference $\Delta$ of KQ and GR peaks as the optimal one: $2.25$ in $25$-node reservoirs \cite{snyderFindingOptimalRandom2012b}, and near $2$ when $\Delta$ is summed over the input fan-out \cite{snyderComputationalCapabilitiesRandom2013}. We test whether the $\kavg$ at which $\Delta$ peaks lies near $\kbest$ across the design grid.

%% file: sections/background.tex
\section{Input Fan-Out in Boolean Reservoirs}\label{sec:background}

In the Boolean reservoirs of Snyder et al.~\cite{snyderFindingOptimalRandom2012b} there is one input node, and the source of each in-edge of a reservoir node is drawn uniformly from the input and reservoir nodes together, one shared pool $(\Iset{\cup}\Rset){\to}\Rset$. Each of the $\kavg\nR$ pool edges has the input node as its source with probability $1/(\nR{+}1)$, so a reservoir has on average about $\kavg$ input edges: the input fan-out grows with $\kavg$. No reservoir node receives the input with probability $p_0 = \bigl(1 - \tfrac{1}{\nR+1}\bigr)^{\kavg\nR} \approx e^{-\kavg}$, at large $\nR$: at $\kavg{=}1$ more than a third of such reservoirs run without input. In this wiring, $\kavg$ thus sets both the number of recurrent edges and the input fan-out.

With the setup of Snyder et al.\ ($\nR{=}25$, one input node) we reproduce their S-curve: KQ (Sect.~\ref{sec:method_kqgr}) rises from low to full rank as $\kavg$ grows. We then change only the input wiring. Drawing the input edges separately from the recurrent edges, as many on average as in the shared pool, changes the curve little at low $\kavg$, so there KQ is limited by how few nodes receive the input, not by the two kinds of edges sharing one pool. A one-to-one mapping from $\redund$ input nodes to $\redund$ reservoir nodes instead fixes the input fan-out at $\redund$ for every $\kavg$, and at $\redund{=}5$ raises KQ at every $\kavg$, at $\kavg{=}1$ from $1.5$ to $4.2$ of $25$. The replication details are in Appendix~\ref{app:snyder}\dashavail{, and every curve and figure of this paper is explorable at \href{\dashbase/furu/}{\dashbase/furu/}}{, and the code behind every curve is available (Sect.~\ref{sec:method_design})}. The rest of the paper uses this mapping over the full design grid (Sects.~\ref{sec:method} and \ref{sec:results_kbest}).

%% file: sections/method.tex
\section{Method}\label{sec:method}

\subsection{Model}\label{sec:method_model}

\begin{figure}[t]
    \centering
    \input{figures/tex/toy_graph}
    \caption{A toy Boolean reservoir with $\nR{=}3$ reservoir nodes, input fan-out $\redund{=}1$ and $\kavg{=}1$. Left: the input bit $u(t)$ is written into the input node $I_1$, which has an edge to $R_1$; the recurrent edges form the cycle $R_1{\to}R_2{\to}R_3{\to}R_1$. Solid edges are fixed, dashed edges are the trained readout $O_1$. Middle: the rules ($'$ marks the next state). $R_1$ receives the input, so its lookup table has one column per combination of $I_1$ and $R_3$; $R_2$ and $R_3$ have one in-neighbour each. Right: the state after each step for $u = 1, 0, 1$ (override injection, one tick, $\mathbf{x}(0) = \mathbf{0}$). At $t{=}1$, for example, $R_1' = I_1 \oplus R_3 = 1 \oplus 0 = 1$: every node reads the state before the update.}
    \label{fig:toy}
\end{figure}
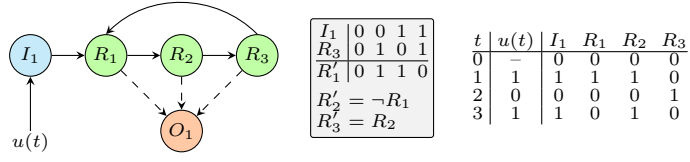

\textbf{Input.} A sample is one stream of bits, $u(t) \in \{0,1\}$ at step $t$. The input fan-out $\redund$ is the number of reservoir nodes that receive each bit.

\textbf{Graph.} The network has input nodes $\Iset$ and reservoir nodes $\Rset$, which form one Boolean node set $\Vset = \Iset \cup \Rset$ with state $\mathbf{x}(t) \in \{0,1\}^{|\Vset|}$, whose reservoir part is $\mathbf{x}_{\Rset}(t)$ (Fig.~\ref{fig:toy}). Two kinds of fixed edges connect them. The input edges $\Iset{\to}\Rset$ are one-to-one: each bit is copied to $\redund \le \nR = 512$ input nodes, each with a single edge to a distinct reservoir node drawn at random (Sect.~\ref{sec:background}), and input nodes receive no edges. The recurrent edges $\Rset{\to}\Rset$ number exactly $\kavg\nR$, so the connectivity $\kavg$ is the mean in-degree of a reservoir node. The wiring mode sets how these edges are spread over the nodes. In the homogeneous mode every node has in-degree $k_i = \kavg$, which requires an integer $\kavg$. In the heterogeneous mode the $\kavg\nR$ edges are drawn without replacement from $\kmax{=}25$ possible in-edges per node, so the in-degree $k_i$ is hypergeometric, at most $\kmax$, with a mean $\kavg$ that can be fractional. At $\kavg \le 8$ no node reaches $\kmax$, and input edges do not count toward it. In both wiring modes each node draws $k_i$ distinct sources uniformly from the reservoir, and exactly a fraction $\selfloop$ of the nodes has a self-loop, which replaces one of its sources. The design grid uses the heterogeneous mode; a second grid in the homogeneous mode, at $\kavg \in \{1, \dots, 8\}$ with one tick, no self-loops and random initialization, tests whether the in-degree distribution moves $\kbest$.

\textbf{Rules.} Each reservoir node $i$ has a rule, a random lookup table (LUT) over its in-neighbours, $f_i : \{0,1\}^{k_i} \to \{0,1\}$. A node that receives the input reads its input node first, so its rule is $f_i : \{0,1\}^{k_i+1} \to \{0,1\}$, as for $R_1$ in Fig.~\ref{fig:toy}. The table entries are drawn once, i.i.d.\ from $\mathrm{Bernoulli}(p)$ with $p = 0.5$, for which the critical connectivity $1/(2p(1-p))$ is $2$ \cite{derridaRandomNetworksAutomata1986}. A reservoir's graph, LUTs and initial state are drawn from one seed (Sect.~\ref{sec:method_design}).

\textbf{Dynamics.} Each step injects the input through the injection map $P$,
\begin{equation}
    [P(\mathbf{x}, u)]_i =
    \begin{cases}
        u, & i \in \Iset \text{ (\emph{override})},\\
        x_i \oplus u, & i \in \Iset \text{ (\emph{xor})},
    \end{cases}
    \label{eq:inject}
\end{equation}
and applies the synchronous update $F$: every node with at least one in-neighbour sets $x_i \leftarrow f_i(\text{its in-neighbours})$, the others keep their value. One step is $\mathbf{x}(t) = F^{\nu}\bigl(P(\mathbf{x}(t{-}1), u(t))\bigr)$, with $\nu \in \{1,2\}$ applications of $F$ per input bit (the ticks). Under \emph{xor} the input node holds the running parity of the stream rather than the current bit. $P$ changes only the input nodes; the input enters the rules of reservoir nodes as one argument. One state per input bit is recorded, after the last application of $F$; unlike CA reservoirs that read every iteration within a step \cite{nicheleReservoirComputingUsing2017}, no readout sees the state between ticks. Each sample starts from the same state $\mathbf{x}(0)$, with input nodes at $0$ and the reservoir part in one of three initialization modes: \emph{zeros}, $\mathbf{x}_{\Rset}(0) = \mathbf{0}$; \emph{random}, $x_i(0) \sim \mathrm{Bernoulli}(0.5)$ i.i.d.; or \emph{random-warmup}, $\mathbf{x}(0) = F^{\nu}(\tilde{\mathbf{x}})$ with $\tilde{\mathbf{x}}$ drawn as in random, that is $\nu$ ticks without input. $\mathbf{x}(0)$ is drawn once per reservoir and is the same for every sample, so it carries no information about the input.

\subsection{Tasks and Readout}\label{sec:method_tasks}

\textbf{Washout.} Every sample starts from the same initial state, which would make a sample's first input bits trivially readable, so each sample begins with $10$ random washout bits of its own, followed by $15$ pattern bits that contain every target window ($T = 25$ steps, $t = 1, \dots, T$). The pattern bits run through all $2^{15}$ patterns once each, so the bits are independent, with maximum variance.

\textbf{Targets.} The targets are computed from one window of $w \in \{3,5\}$ pattern bits at delay $d \in \{5,10\}$ from the end of the stream: the $w$ bits at $t = T{-}d{-}w{+}1, \dots, T{-}d$.
\begin{equation}
    y^{\mathrm{maj}} = \Bigl[\,2\!\!\sum_{t=T-d-w+1}^{T-d}\!\! u(t) > w\Bigr], \qquad
    y^{\mathrm{par}} = \bigoplus_{t=T-d-w+1}^{T-d} u(t),
    \label{eq:targets}
\end{equation}
Majority asks whether most bits of the window are $1$; parity is the XOR of all $w$ bits. Both tasks follow Snyder et al.~\cite{snyderFindingOptimalRandom2012b}, who call majority density classification, as in CA \cite{mitchellRevisitingEdgeChaos1993}. A \emph{task} is majority or parity, a \emph{task variant} is one task at one window, delay and readout ($2^4 = 16$ in all), and \emph{target} is reserved for the $y$ of Eq.~\eqref{eq:targets}. All task variants read the same reservoir run.

\textbf{Split.} The $2^{15}$ samples, one per pattern, are shuffled and split $50/30/20$ into training, development and test sets, so no pattern is in two sets. Each target value holds for exactly half of the patterns, since flipping one window bit flips the parity and flipping all of them flips the majority of an odd window, so the split needs no stratification. The readout trains on the training set, and every reported accuracy is on the development set unless marked as test. The test set only checks that the choices of the analysis did not fit the development samples: the mean $|$dev$-$test$|$ difference is $0.5$--$0.8\,\%$ per task variant, and at most $0.9\,\%$ at any $\kavg$, close to the sampling noise of the split ($0.64\,\%$).

\textbf{Readout.} A linear readout reads the reservoir nodes only, at the steps $\mathcal{S}$: $\mathcal{S} = \{T\}$ for the \emph{final-state} readout and $\mathcal{S} = \{11, \dots, T\}$, the steps after the washout, for the \emph{summed} readout. Each task variant has its own logistic regression
\begin{equation}
    \hat{p} = \operatorname{sigmoid}\Big(\mathbf{w} \cdot \sum_{t \in \mathcal{S}} \mathbf{x}_{\Rset}(t) + \beta\Big),
    \label{eq:readout}
\end{equation}
with $\mathbf{x}_{\Rset}$ the reservoir part of the state, fitted by Newton--IRLS with an $L_2$ penalty of $10^{-6}$ relative to the feature scale, only to keep separable and rank-deficient fits well posed. The readout predicts $1$ when $\hat{p} > 0.5$ and is evaluated by its development-set accuracy.

\subsection{Kernel Quality and Generalization Rank}\label{sec:method_kqgr}

    KQ and GR \cite{legensteinEdgeChaosPrediction2007}, and their difference $\Delta$, follow Snyder et al.~\cite{snyderFindingOptimalRandom2012b}, except that $\Delta$ is signed (theirs is $|\mathrm{KQ} - \mathrm{GR}|$) and each stream begins with the washout. A probe set $\mathcal{U} = \{\mathbf{u}^{(1)}, \dots, \mathbf{u}^{(m)}\}$ holds $m = \nR = 512$ streams built like samples, so that the state matrix is square, as in Snyder et al. The readout input $\sum_{t \in \mathcal{S}} \mathbf{x}^{(n)}_{\Rset}(t)$ of stream $\mathbf{u}^{(n)}$ forms row $n$ of the state matrix $X(\mathcal{U}) \in \mathbb{R}^{m \times \nR}$, so KQ, GR and $\Delta$ are measured separately for each readout: on the final state, as in Snyder et al., or on the summed states. Then
\begin{equation}
    \mathrm{KQ} = \operatorname{rank} X(\mathcal{U}), \qquad
    \mathrm{GR} = \operatorname{rank} X(\mathcal{U}_{\tau}), \qquad
    \Delta = \mathrm{KQ} - \mathrm{GR},
    \label{eq:kqgr}
\end{equation}
with the numerical rank over $\mathbb{R}$. $\mathcal{U}_{\tau}$ is $\mathcal{U}$ with the last $\tau{=}5$ bits of every stream replaced by those of one reference stream, and its streams are drawn so that they stay distinct. This construction is one of $12$, chosen on the selection run (Sect.~\ref{sec:method_design}); Appendix~\ref{app:probe} gives the others and the further results of KQ, GR and $\Delta$. Every reservoir is measured on the same streams.

\subsection{Experiments}\label{sec:method_design}

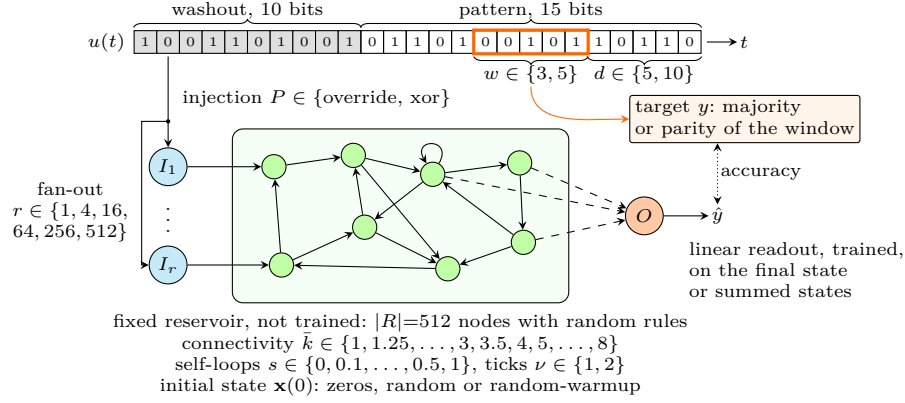
\begin{figure}[tb]
\centering
\input{figures/tex/setup_overview}
\caption{The setup and the design grid. A sample is a stream of $25$ bits, the first $10$ washout; the target is the majority or the parity of the last $w$ bits, shifted back by $d$. The figure lists the levels of each design factor (Table~\ref{tab:factors} in the appendix) and the windows and delays of the task variants; only the linear readout is trained.}
\label{fig:setup}
\end{figure}

\textbf{Design grid.} A design takes one level of each design factor in Fig.~\ref{fig:setup}, and the grid is full factorial: $504$ combinations of input fan-out, injection, ticks, initialization and self-loops, each at $15$ values of $\kavg$, give $7{,}560$ designs.

\textbf{Reservoirs.} Each design is built as $20$ reservoirs, each with its own random wiring and rules, seeded by the design and the reservoir's index, so no two designs share a reservoir. Every reservoir is evaluated on all $16$ task variants (Sect.~\ref{sec:method_tasks}) on the same $2^{15}$ patterns and split, and measured by KQ and GR, so task variants are always compared on the same reservoirs. A design's \emph{design accuracy} on a task variant, or \emph{accuracy} for short, is the median over its $20$ reservoirs. Every choice of the analysis (the construction of $\mathcal{U}_{\tau}$, each task variant's best design and the $\kavg$ at which self-loops are compared, below) was made on a selection run of the same grid with $12$ other reservoirs per design, so the choices and the reported numbers come from different reservoirs. Held-out patterns alone would not suffice, since a design's accuracy varies far more between its reservoirs than between pattern sets (Appendix~\ref{app:stats}).

\textbf{Floors.} The \emph{floor} of a task variant is the best accuracy without a reservoir, predicting from the last input bit (final-state readout) or from the count of ones among the pattern bits (summed readout). The floor is $66$--$71\,\%$ for summed majority and chance ($50$--$51.5\,\%$) otherwise.

\textbf{Operating point.} The \emph{operating point} of a task variant is the $\redund$, $P$ and $\kavg$ of its best design. Ticks, initialization and self-loops vary freely there, giving $42$ designs per task variant. $\kavg$ is fixed as well because with only $\redund$ and $P$ fixed many designs are at the floor, where the other factors show no difference. When ticks or initialization are compared, self-loops are held equal, as $\kavg$ is, since a self-loop replaces one of its node's $k_i$ sources and so changes the number of edges between distinct nodes. Self-loops are compared on each combination of the other factors, placed at the $\kavg$ where it is most accurate on the selection run (its own $\kbest$), against the same combination without self-loops, placed likewise.

\textbf{Analysis.} Accuracies are in $\%$, and differences between them are absolute. The analysis makes two choices. First, input fan-out and injection are compared by the best design over the whole grid, and $\kavg$ by mean curves (design accuracy averaged over all designs at one fan-out and $\kavg$ and over a task's four variants, with the optimal connectivity $\kbest$ at a curve's peak). Every other factor is compared at the operating point, since fan-out and $\kavg$ determine whether a design rises above its floor, and over the whole grid their effects are larger than those of the other factors. Second, designs are always compared by design accuracy, the median defined above, since reservoirs of the same design differ widely; a design is never judged by its best reservoir. Appendix~\ref{app:stats} describes how each interval was computed, and a script for every number in the paper is at \newline\codelink.

%% file: figures/tex/toy_graph.tex
\begin{tikzpicture}[
    every node/.style={minimum size=5.5mm, inner sep=0.5pt, font=\scriptsize},
    input node/.style={draw, circle, fill=cyan!20},
    reservoir node/.style={draw, circle, fill=green!60!lime!35},
    output node/.style={draw, circle, fill=orange!80!red!35},
    edge/.style={->, >=stealth},
    readout/.style={->, >=stealth, dashed},
    tab/.style={minimum size=0pt, inner sep=0pt, font=\scriptsize},
    ]
    \node[input node]     (I1) at (0.0, 0) {$I_1$};
    \node[reservoir node] (R1) at (1.0, 0) {$R_1$};
    \node[reservoir node] (R2) at (2.0, 0) {$R_2$};
    \node[reservoir node] (R3) at (3.0, 0) {$R_3$};
    \node[output node]    (O1) at (2.0,-1.0) {$O_1$};

    \draw[edge] (0.0,-1.0) node[below, minimum size=0pt] {$u(t)$} -- (I1);
    \draw[edge] (I1) -- (R1);
    \draw[edge] (R1) -- (R2);
    \draw[edge] (R2) -- (R3);
    \draw[edge] (R3.north) to[out=135, in=45] (R1.north);

    \draw[readout] (R1) -- (O1);
    \draw[readout] (R2) -- (O1);
    \draw[readout] (R3) -- (O1);

    \node[tab, draw, rounded corners=1pt, fill=gray!10, inner sep=2pt, anchor=west, align=left] (rules) at (3.7, -0.3) {%
        \renewcommand{\arraystretch}{0.85}\setlength{\tabcolsep}{2pt}%
        \begin{tabular}{@{}c|cccc@{}}
        $I_1$  & 0 & 0 & 1 & 1\\
        $R_3$  & 0 & 1 & 0 & 1\\ \hline
        $R_1'$ & 0 & 1 & 1 & 0
        \end{tabular}\\[2pt]
        $R_2' = \lnot R_1$\\
        $R_3' = R_2$};

    \node[tab, anchor=west] at ([xshift=5mm]rules.east) {%
        \renewcommand{\arraystretch}{0.85}\setlength{\tabcolsep}{2.5pt}%
        \begin{tabular}{@{}c|c|cccc@{}}
        $t$ & $u(t)$ & $I_1$ & $R_1$ & $R_2$ & $R_3$\\ \hline
        0 & -- & 0 & 0 & 0 & 0\\
        1 & 1  & 1 & 1 & 1 & 0\\
        2 & 0  & 0 & 0 & 0 & 1\\
        3 & 1  & 1 & 0 & 1 & 0
        \end{tabular}};
\end{tikzpicture}

%% file: figures/tex/setup_overview.tex
\begin{tikzpicture}[
    every node/.style={font=\scriptsize},
    bit/.style={draw, minimum size=3mm, inner sep=0pt, font=\tiny},
    wash/.style={bit, fill=gray!25},
    input node/.style={draw, circle, fill=cyan!20, minimum size=5mm, inner sep=0pt},
    reservoir node/.style={draw, circle, fill=green!60!lime!35, minimum size=3.2mm, inner sep=0pt},
    output node/.style={draw, circle, fill=orange!80!red!35, minimum size=5mm, inner sep=0pt},
    edge/.style={->, >=stealth},
    readout/.style={->, >=stealth, dashed},
    note/.style={align=center, inner sep=1pt},
    ]
    \foreach \b [count=\i from 0] in {1,0,0,1,1,0,1,0,0,1, 0,1,1,0,1,0,0,1,0,1,1,0,1,1,0} {
        \ifnum\i<10
            \node[wash] (u\i) at (0.3*\i+0.6, 3.1) {\b};
        \else
            \node[bit] (u\i) at (0.3*\i+0.6, 3.1) {\b};
        \fi
    }
    \draw[very thick, orange!80!red] (u15.north west) rectangle (u19.south east);
    \draw[decorate, decoration={brace, amplitude=3pt}] ([yshift=1pt]u0.north west) -- ([yshift=1pt]u9.north east)
        node[midway, above=2pt, note] {washout, $10$ bits};
    \draw[decorate, decoration={brace, amplitude=3pt}] ([yshift=1pt]u10.north west) -- ([yshift=1pt]u24.north east)
        node[midway, above=2pt, note] {pattern, $15$ bits};
    \draw[decorate, decoration={brace, amplitude=3pt, mirror}] ([yshift=-1pt]u15.south west) -- ([yshift=-1pt]u19.south east)
        node[midway, below=2pt, note] (win) {$w \in \{3,5\}$};
    \draw[decorate, decoration={brace, amplitude=3pt, mirror}] ([yshift=-1pt]u20.south west) -- ([yshift=-1pt]u24.south east)
        node[midway, below=2pt, note] {$d \in \{5,10\}$};
    \draw[->, >=stealth] ([xshift=2pt]u24.east) -- ++(0.4, 0) node[right, inner sep=1pt] {$t$};
    \node[anchor=east, note] at ([xshift=-2pt]u0.west) {$u(t)$};

    \node[draw, rounded corners=1pt, fill=orange!8, align=left, anchor=north west, inner sep=2pt] (target)
        at (7.0, 2.4) {target $y$: majority\\or parity of the window};
    \draw[->, >=stealth, orange!80!red] (win.south) to[out=-90, in=180] (target.west);

    \node[input node] (I1) at (0.9, 1.45) {$I_1$};
    \node (dots) at (0.9, 0.85) {$\vdots$};
    \node[input node] (Ir) at (0.9, 0.15) {$I_r$};
    \node[note, anchor=east] at (0.42, 0.85) {fan-out\\$r \in \{1, 4, 16,$\\$64, 256, 512\}$};
    \draw ([yshift=-1pt]u1.south) -- (0.9, 2.05) coordinate (split);
    \fill (split) circle (0.8pt);
    \draw[->, >=stealth] (split) -- (I1.north);
    \draw[->, >=stealth] (split) -- (0.55, 2.05) -- (0.55, 0.15) -- (Ir.west);
    \node[note, anchor=west] at (1.05, 2.35) {injection $P \in \{$override, xor$\}$};

    \draw[rounded corners=4pt, fill=green!4] (1.75, -0.35) rectangle (6.2, 1.95);
    \node[reservoir node] (a) at (2.3, 1.45) {};
    \node[reservoir node] (b) at (3.35, 1.6) {};
    \node[reservoir node] (c) at (4.4, 1.35) {};
    \node[reservoir node] (d) at (5.55, 1.5) {};
    \node[reservoir node] (e) at (2.4, 0.15) {};
    \node[reservoir node] (f) at (3.5, 0.65) {};
    \node[reservoir node] (g) at (4.6, 0.1) {};
    \node[reservoir node] (h) at (5.6, 0.45) {};
    \foreach \s/\t in {a/b, b/c, c/d, d/h, h/g, g/e, e/a, f/b, c/f, f/g, e/f, b/g, h/c}
        \draw[edge] (\s) -- (\t);
    \draw[edge] (c) to[out=60, in=120, looseness=6] (c);
    \draw[edge] (I1) -- (a);
    \draw[edge] (Ir) -- (e);

    \node[output node] (O) at (7.2, 0.8) {$O$};
    \foreach \s in {c, d, h}
        \draw[readout] (\s) -- (O);
    \draw[->, >=stealth] (O) -- (8.05, 0.8) node[right, inner sep=1pt] (yhat) {$\hat{y}$};
    \draw[<->, >=stealth, densely dotted] (yhat.north) -- (yhat.north |- target.south) node[midway, right, note] {accuracy};

    \node[note, anchor=north] at (3.975, -0.45) {fixed reservoir, not trained: $|R|{=}512$ nodes with random rules\\
        connectivity $\kavg \in \{1, 1.25, \dots, 3, 3.5, 4, 5, \dots, 8\}$\\
        self-loops $s \in \{0, 0.1, \dots, 0.5, 1\}$, ticks $\nu \in \{1, 2\}$\\
        initial state $\mathbf{x}(0)$: zeros, random or random-warmup};
    \node[note, anchor=north west, align=left] at (7.75, 0.5) {linear readout, trained,\\on the final state\\or summed states};
\end{tikzpicture}

%% file: sections/results.tex
\section{Results}\label{sec:results}

\subsection{Accuracy, Input Fan-Out and Injection}\label{sec:results_fanout}

\begin{figure}[!ht]
    \centering
    \dashfig{/furu/analysis/}{design_acc_by_r_final_state}{\includegraphics[width=0.49\linewidth]{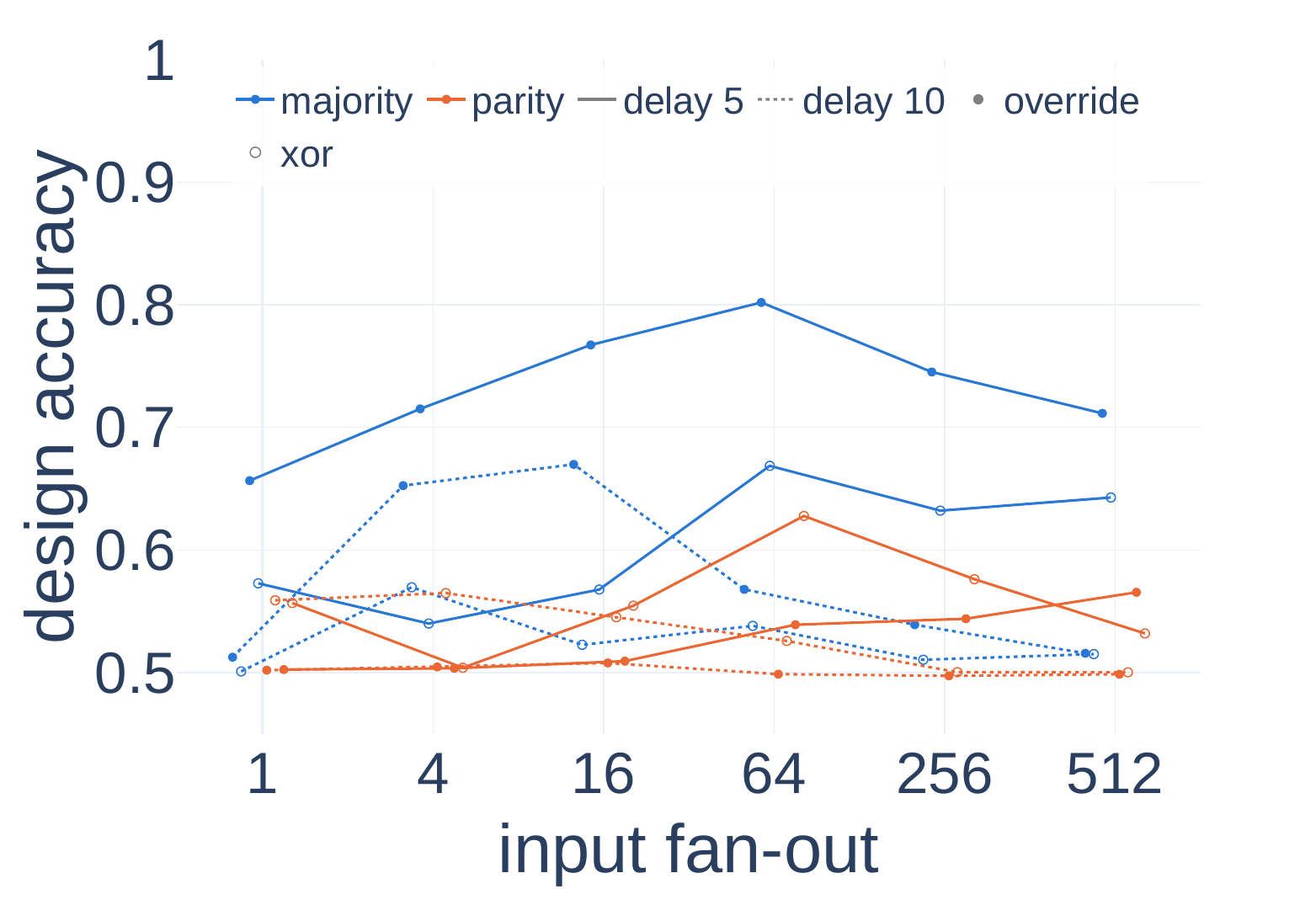}}\hfill
    \dashfig{/furu/analysis/}{design_acc_by_r_summed}{\includegraphics[width=0.49\linewidth]{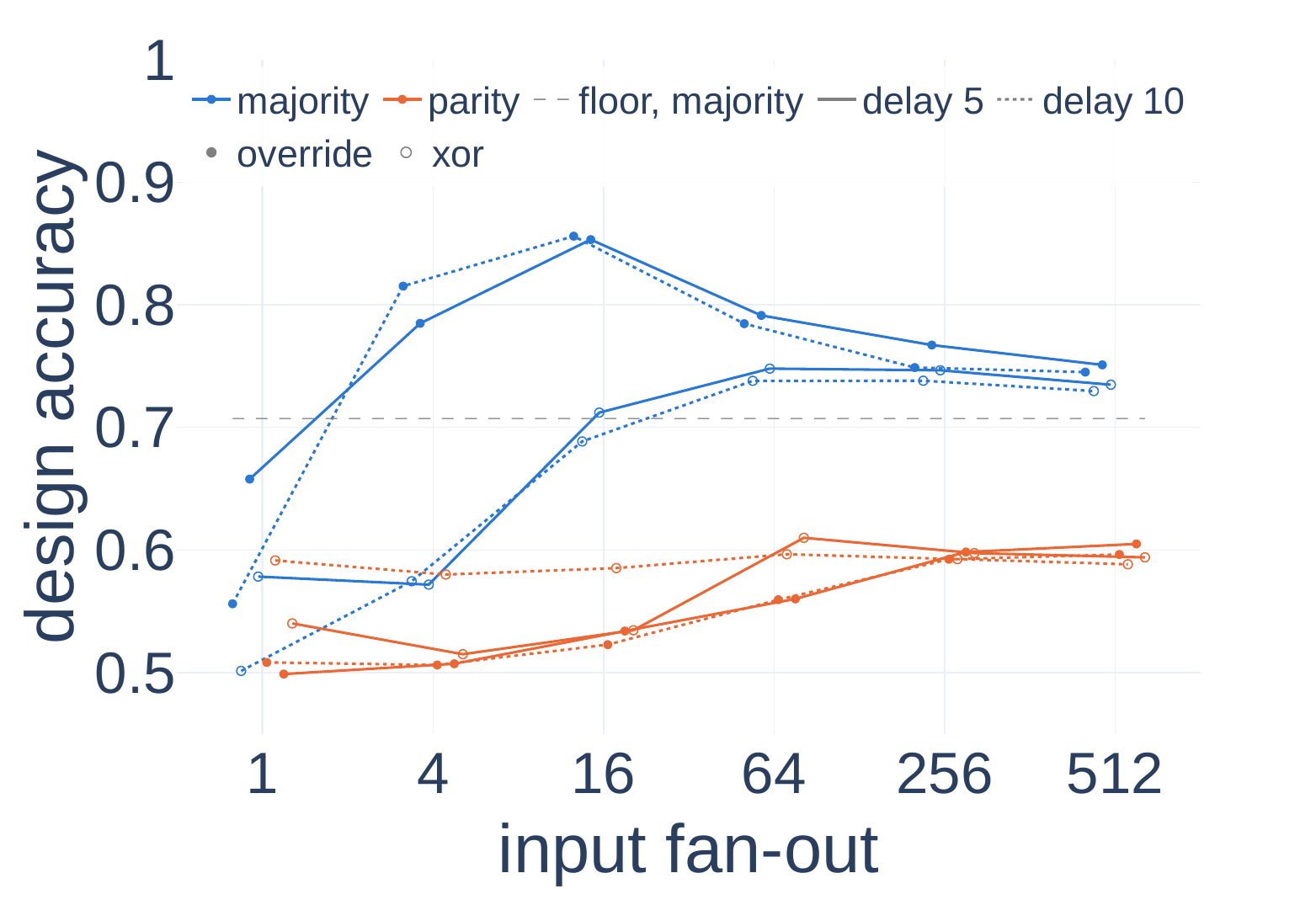}}
    \caption{Design accuracy of the best design at each input fan-out and injection ($w{=}5$), final-state readout (left) and summed (right). The best design over $\kavg$, ticks, initialization and self-loops is chosen per task variant on the selection run (Sect.~\ref{sec:method_design}). Colour: task; line style: delay (solid $d{=}5$, dotted $d{=}10$); marker: injection (filled override, open xor). Dashed grey line: the floor of summed majority ($70.7\,\%$ at both delays).}
    \label{fig:fanout}
\end{figure}

Every task variant's best design is $9$--$41\,\%$ more accurate than its floor (Sect.~\ref{sec:method_design}), except final-state parity at $d{=}10$ ($6\,\%$). At $w{=}5$ (Fig.~\ref{fig:fanout}), on the final-state readout, majority is most accurate at fan-out $64$ at $d{=}5$ ($80\,\%$) and at $16$ at $d{=}10$ ($67\,\%$), and parity at $64$ at $d{=}5$ ($63\,\%$, against at most $56\,\%$ at lower fan-out); parity at $d{=}10$ stays at $50$--$57\,\%$ at every fan-out. On the summed readout, majority is most accurate at fan-out $16$ ($85$--$86\,\%$), and parity at $64$ at $d{=}5$ ($61\,\%$); parity at $d{=}10$ stays at $58$--$60\,\%$. Beyond its best fan-out every majority variant is less accurate, by $9$--$15\,\%$ at fan-out $512$. At $w{=}3$ accuracy is higher, at the same best fan-outs except on summed parity at $d{=}10$ ($256$) (up to $91\,\%$ on majority and $72\,\%$ on parity).

Between the best override design and the best xor design over all fan-outs, on the final-state readout override is more accurate on majority, by $8$--$17\,\%$, and xor on parity at $d{=}5$, by $6\,\%$ at $w{=}5$ and $1\,\%$ at $w{=}3$. On the summed readout override is more accurate on majority, by $10$--$14\,\%$, and on parity override and xor differ by at most $2\,\%$.

\subsection{The Optimal Connectivity}\label{sec:results_kbest}

\begin{figure}[tb]
    \centering
    \dashfig{/furu/analysis/}{design_k_best_by_r}{\includegraphics[width=\linewidth]{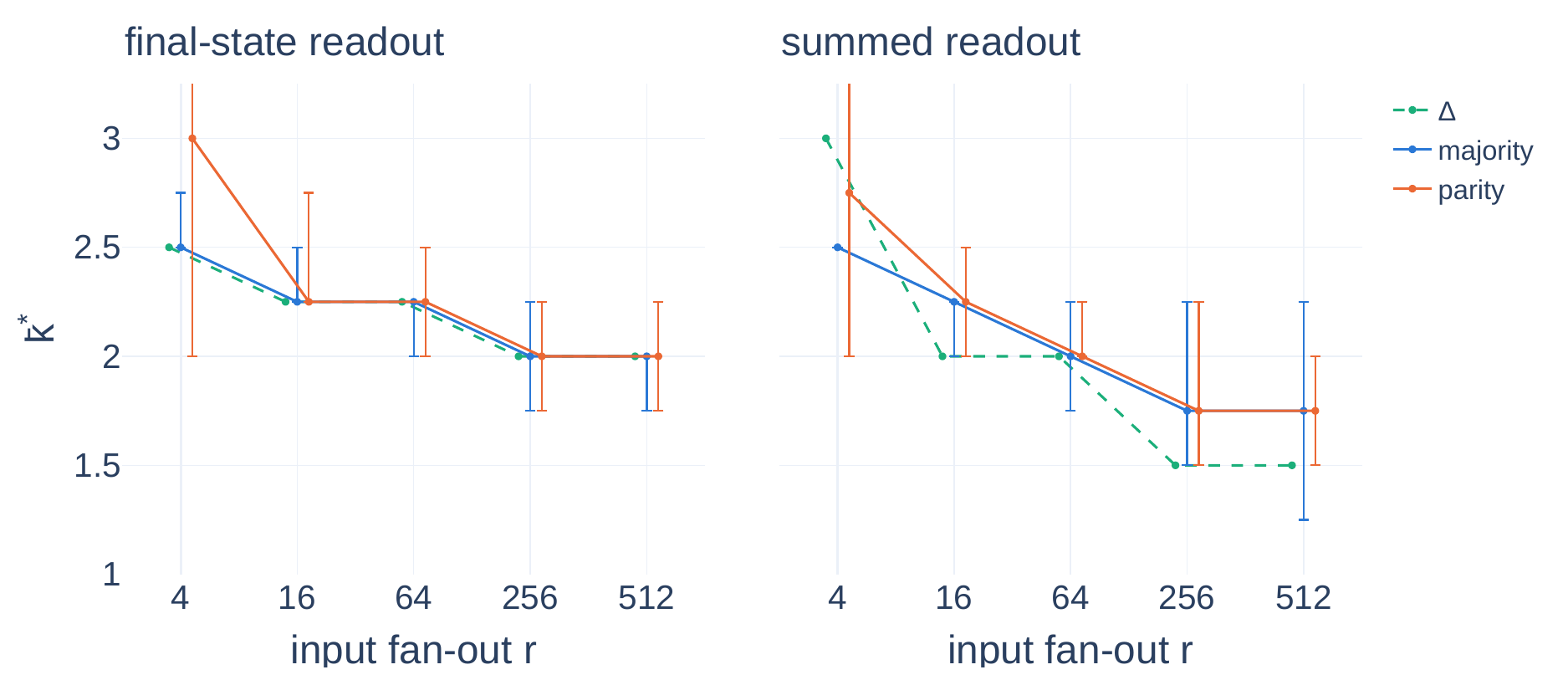}}
    \caption{The optimal connectivity $\kbest$ against input fan-out (categorical axis), final-state readout (left) and summed (right). Solid lines: $\kbest$ of each task, from the mean design-accuracy curve over all designs at that fan-out and $\kavg$ and over the task's four variants (windows and delays); bars span the $\kavg$ within $1\,\%$ of the accuracy at $\kbest$. Dashed lines: the $\kavg$ at the peak of the mean $\Delta$ curve over the same designs; $\Delta$ does not depend on the task, so each readout has one. Fan-out $1$ is left out: its mean curves are within $1\,\%$ of their peak from $\kavg{=}3.5$--$5$ up to $8$. At fan-out $4$ the parity bars run to $\kavg{=}8$, beyond the axis.}
    \label{fig:kbest}
\end{figure}

In Fig.~\ref{fig:kbest}, $\kbest$ of the majority task falls as the fan-out grows, on both readouts: from $2.5$ at fan-out $4$, above the critical value $\kavg{=}2$ of RBNs, to $2$ (final-state) and $1.75$ (summed) at fan-out $512$. Its $95\,\%$ interval spans at most one grid step ($0.25$), and the median of each combination's own $\kbest$ falls in the same way (Appendix~\ref{app:stats}). On the majority task, $\kavg{=}2$ is $3.5$--$4.5\,\%$ less accurate than $\kbest$ at fan-out $4$, $0.8$--$2.6\,\%$ at $16$, and at most $0.2\,\%$ from $64$ up. Parity's $\kbest$ equals the majority task's from fan-out $16$ up. At fan-out $4$ parity's mean curves stay within $2\,\%$ of chance, so their $\kbest$ says little: from $\kavg{=}2$ to $8$ they stay within $1\,\%$ of their maximum.

The dashed lines in Fig.~\ref{fig:kbest} mark the $\kavg$ at which $\Delta$ peaks, computed without a trained readout (Sect.~\ref{sec:method_kqgr}). On the final-state readout this $\kavg$ falls from $2.5$ at fan-out $4$ to $2$ at $512$; it is at most $0.4\,\%$ less accurate than $\kbest$ at every fan-out, and more accurate than $\kavg{=}2$ at fan-out $4$ and $16$. On the summed readout the $\kavg$ at $\Delta$'s peak falls from $3$ to $1.5$; on majority at fan-out $4$ it is $3.1\,\%$ less accurate than $\kbest$, and $\kavg{=}2$ is $3.5\,\%$ less accurate. From fan-out $16$ up, the $\kavg$ at $\Delta$'s peak and $\kavg{=}2$ are both at most $0.8\,\%$ less accurate than $\kbest$.

The task and the delay move $\kbest$ beyond its interval only where the best design is at most $58\,\%$ accurate. We compare pairs of task variants that differ only in the task (task pairs) or only in the delay (delay pairs), on the same reservoirs at each fan-out from $4$ up (Appendix~\ref{app:stats}), leaving out three pairs of each kind, all with final-state parity at $d{=}10$ ($50\,\%$). The two variants of a pair have the same $\kbest$ in $26$ of the remaining $37$ task pairs and in $24$ of $37$ delay pairs, and a longer delay never lowers $\kbest$. All $8$ pairs whose $95\,\%$ interval of the difference excludes $0$ involve such a variant.

Injection, ticks and initialization move $\kbest$ by at most one grid step in $59$ of the $60$ combinations of factor, readout, task and fan-out. Self-loops move $\kbest$ in one direction: from $s{=}0$ to $s{=}1$, $\kbest$ rises in all $20$ combinations of readout, task and fan-out, by $0.25$--$0.75$, less than the one edge from another node that each self-loop replaces. The $\kavg$ at $\Delta$'s peak rises with self-loops as well: at $s{=}1$ it lies within $0.5$ of $\kbest$ in $95\,\%$ of these combinations, and $\kavg{=}2$ in $50\,\%$.

Homogeneous wiring, every node at in-degree exactly $\kavg$, gives the same $\kbest$ as heterogeneous wiring on the integer grid $\kavg \in \{1, \dots, 8\}$ in $75\,\%$ of the $96$ combinations of task variant and fan-out, and within $1$ in $91\,\%$. The best homogeneous design is at least as accurate, and more accurate at fan-out $64$ and $256$, by $2.7$ and $2.0\,\%$ on average, with $95\,\%$ intervals above $0$ (Appendix~\ref{app:stats}).

\begin{table}[tb]
\centering
\small
\caption{Effect of ticks and self-loops at the operating point: the median difference in design accuracy between designs that differ only in that factor, relative to one tick and no self-loops, in $\%$. The self-loop rows place each combination at its own $\kbest$ (Sect.~\ref{sec:method_design}). Each column takes the pairs from its four task variants ($84$ pairs for ticks, $24$ per self-loop level). Brackets: the interquartile range over the pairs.}
\label{tab:choices}
\begin{tabular}{lcccc}
\toprule
 & final-state & final-state & summed & summed \\
 & majority & parity & majority & parity \\
\midrule
ticks $\nu{=}2$       & $-10.6$ & $-4.8$ & $-6.2$ & $-3.3$ \\
                      & {\scriptsize$[-14.2, -5.9]$} & {\scriptsize$[-6.4, -3.4]$} & {\scriptsize$[-9.7, -2.7]$} & {\scriptsize$[-4.1, -1.1]$} \\[3pt]
self-loops $s{=}0.1$  & $-2.2$ & $-0.1$ & $-0.9$ & $0.0$ \\
                      & {\scriptsize$[-3.9, +0.4]$} & {\scriptsize$[-2.0, +1.2]$} & {\scriptsize$[-2.3, +0.5]$} & {\scriptsize$[-1.2, +0.7]$} \\[3pt]
self-loops $s{=}1.0$  & $-9.8$ & $-1.4$ & $-6.3$ & $-3.4$ \\
                      & {\scriptsize$[-15.5, -3.7]$} & {\scriptsize$[-7.2, +0.5]$} & {\scriptsize$[-10.4, -0.4]$} & {\scriptsize$[-5.0, -0.1]$} \\
\bottomrule
\end{tabular}
\end{table}

\subsection{Other Design Factors and Variation Between Reservoirs}\label{sec:results_choices}

At the operating point (Table~\ref{tab:choices}), a second tick lowers accuracy by $3$--$11\,\%$ on every readout and task. Self-loops lower accuracy, more on the majority task than on parity; with self-loops on a tenth of the nodes ($s{=}0.1$), no median decrease exceeds $2.2\,\%$. A Beta regression of single-reservoir accuracy finds a larger decrease at $s{=}1$ than at $s{=}0.1$ on every readout and task (Appendix~\ref{app:stats}). The initialization changes design accuracy by at most $0.5\,\%$.

Reservoirs of the same design differ in their random wiring and rules. On the majority task these differences account for $17$--$45\,\%$ of the variance of single-reservoir accuracy over the grid, and on parity for $16$--$65\,\%$, most on final-state parity at $d{=}10$ ($63$--$65\,\%$). In each task variant's best design, the best of the $20$ reservoirs is more accurate than the median one by $6$--$28\,\%$, more than the median effect of any design change at the operating point. Single reservoirs reach $100\,\%$ on every task variant (at least $99.86\,\%$ on test), on parity even in designs whose median is about $51\,\%$.

%% file: sections/discussion.tex
\section{Discussion}\label{sec:discussion}

\textbf{What moves the optimal connectivity.} Against our expectation, $\kbest$ moves only with the fan-out and self-loops; the task, the delay, injection, ticks and initialization hardly move it. Snyder et al.~\cite{snyderFindingOptimalRandom2012b} find parity's $\kbest$ at or above that of the majority task and a lower $\kbest$ at longer delays; here neither appears where accuracy rises above chance, but their reservoirs use a shared pool, and their readout is trained online and evaluated at every step. With a self-loop on every node, each node has one edge from another node fewer, yet $\kbest$ rises by only $0.25$--$0.75$, not by $1$. Either a node's own previous state partly replaces the missing edge, or self-loops move the transition between order and chaos, since a self-loop carries a change in its node's state to no other node; neither explanation was tested.

\textbf{Connectivity and input.} The critical value $\kavg{=}2$ of RBNs \cite{kauffmanMetabolicStabilityEpigenesis1969,derridaRandomNetworksAutomata1986} is close to the optimal connectivity only when many nodes receive the input (fan-out $64$ and up); with few input nodes, reservoirs are most accurate above $\kavg{=}2$, where the undriven network is chaotic. The classical critical point applies here only approximately: nodes that receive the input have $\kavg{+}1$ inputs on average, one of them the injected bit, and a driven network is not a closed system. One untested interpretation: the more nodes the input reaches, the more changes it injects, and the lower $\kavg$ must be to keep the network at the same point between order and chaos. Since $\kbest$ depends on the fan-out, studies with different fan-outs can find different values of $\kbest$, which may explain part of why earlier optima range from $\kavg{=}1$ to $8$. Two earlier findings agree in direction with the fall of $\kbest$ as the fan-out grows: Echlin et al.~\cite{echlinFlexibilityBooleanNetwork2018} find $\kavg{=}1$ best in large reservoirs with many input nodes, and Snyder et al.~\cite{snyderComputationalCapabilitiesRandom2013} find that their more chaotic reservoirs reach their best accuracy with a smaller fan-out. In a shared pool, where $\kavg$ also sets the fan-out, an optimum mixes the effects of $\kavg$ and fan-out \cite{snyderFindingOptimalRandom2012b}, and $\Delta$ summed over the fan-out cannot show how $\kbest$ changes with fan-out \cite{snyderComputationalCapabilitiesRandom2013}.

\textbf{Estimating the optimal connectivity.} Two estimates of $\kbest$ need no trained readout: $\kavg{=}2$, and the $\kavg$ at $\Delta$'s peak, which falls with fan-out and rises with self-loops, as $\kbest$ does. On the final-state readout, the $\kavg$ at $\Delta$'s peak is the better estimate at fan-out $4$ and $16$. On the summed readout, at fan-out $4$, the $\kavg$ at $\Delta$'s peak and $\kavg{=}2$ are both more than $3\,\%$ less accurate than $\kbest$ on the majority task; from fan-out $16$ up, each is within $0.8\,\%$. Neither estimate replaces training a readout.

\textbf{Accuracy.} The explanations in this paragraph fit the results but were not tested. The task changes how accurate a design can be: final-state parity at $d{=}10$ stays at $50$--$57\,\%$ at every fan-out, in line with the scale from memory to computation (Sect.~\ref{sec:introduction}): the majority of a window can still be guessed when some of its bits are lost over the later input, its parity cannot. Under xor the input node holds the running parity of the stream (Sect.~\ref{sec:method_model}), so a window's parity is the XOR of only two past values of the input node, which may explain why xor is the better injection on final-state parity at $d{=}5$ and override on majority. A second tick passes an input bit's effect through two rules before the next bit arrives, so the state keeps fewer past input bits, which fits the lower accuracy on every readout and task. The initialization makes no difference, as expected if the ten washout bits remove the effect of the initial state.

\textbf{Limits.} This study has one reservoir size, one input distribution and two tasks at short delays; self-loops that add an edge instead of replacing one, and a readout of every tick, are untested. Our intervals, uncorrected for multiple comparisons, cover only the variation between reservoirs of the same design, so the results hold for the explored grids. Two questions remain open: why $\kbest$ is near the critical value at high fan-out, and what sets apart the single reservoirs that reach $100\,\%$ where their design's median is about $51\,\%$.

%% file: sections/conclusion.tex
\section{Conclusion}\label{sec:conclusion}
In the reservoirs studied here, the optimal connectivity falls as the input fan-out grows and rises with self-loops, but hardly depends on the task or on the other design factors. The critical value $\kavg{=}2$ is close to optimal only when many nodes receive the input (fan-out $64$ and up). When few do (fan-out $4$), majority is most accurate at $\kavg{=}2.5$ and up to $4.5\,\%$ less accurate at $\kavg{=}2$, and parity shows no clear optimum. The connectivity should therefore be chosen together with the input fan-out and self-loops. The $\kavg$ at $\Delta$'s peak estimates the optimal connectivity on the final-state readout but not on the summed one.

Both tasks are solved at least $9\,\%$ above their floors, except final-state parity at the longer delay. Majority is most accurate at an intermediate fan-out and with override injection, parity at a high fan-out; one state update per input bit is better than two on every task, self-loops on many nodes lower accuracy, and the initialization does not matter. Single reservoirs of one design differ more than a change of ticks, self-loops or initialization changes their median, so a design, and its optimal connectivity, should be judged by the median of many reservoirs.

%% file: sections/appendix.tex
\appendix

\clearpage
\section{Supplementary Material}\label{app:appendix}

\subsection{Design Factors}\label{app:factors}

\begin{table}[h]
\centering
\footnotesize
\setlength{\tabcolsep}{3pt}
\caption{The design factors of the grid (Fig.~\ref{fig:setup}), their values, and what each controls. Every factor is defined in Sect.~\ref{sec:method_model}.}
\label{tab:factors}
\begin{tabular}{@{}ll>{\raggedright\arraybackslash}p{4.4cm}>{\raggedright\arraybackslash}p{3.2cm}@{}}
\toprule
factor & symbol & values & controls \\
\midrule
input fan-out & $\redund$ & $1$, $4$, $16$, $64$, $256$, $512$ & $|\Iset{\to}\Rset|$ \\
injection & $P$ & override, xor & how $u(t)$ enters the input nodes (Eq.~\eqref{eq:inject}) \\
ticks & $\nticks$ & $1$, $2$ & applications of $F$ per input bit \\
initialization & $\mathbf{x}_{\Rset}(0)$ & zeros, random, random-warmup & initial state vector \\
self-loops & $\selfloop$ & $0$--$0.5$ by $0.1$, then $1$ & fraction of nodes with a self-loop \\
connectivity & $\kavg$ & $1$--$3$ by $0.25$; $3.5$, $4$, $5$, $6$, $7$, $8$ & $\sum_i k_i/|\Rset|$ \\
wiring mode & & heterogeneous; homogeneous in a second grid (Sect.~\ref{sec:results_kbest}) & distribution of $k_i$ \\
\bottomrule
\end{tabular}
\end{table}

\subsection{Snyder Replication}\label{app:snyder}

\begin{figure}[h]
\centering
\dashfig{/furu/figure1/}{snyder_kq_vs_k_avg_variants}{\includegraphics[width=\linewidth]{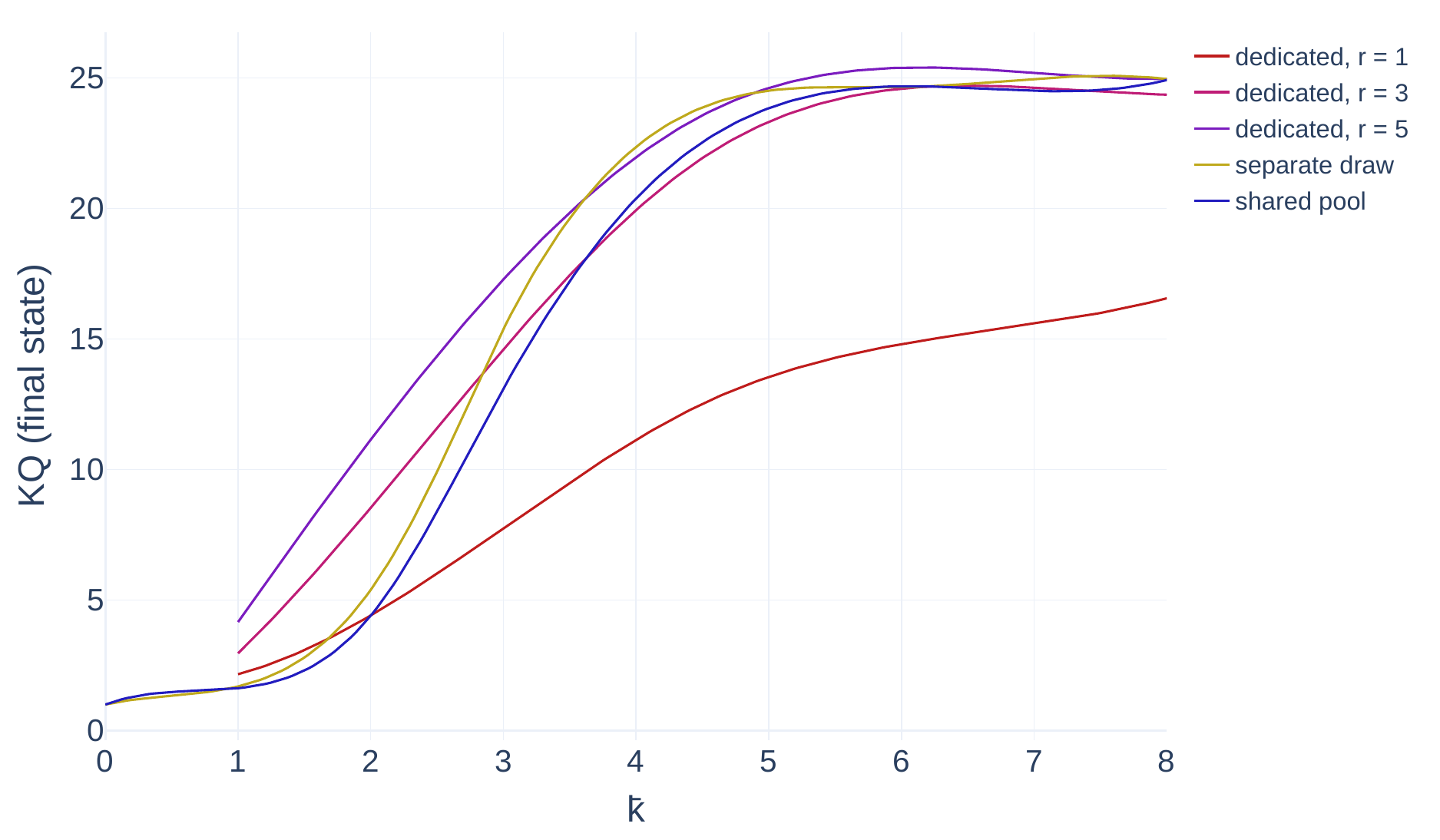}}
\caption{Replication of Snyder et al.'s kernel-quality curve: the rank of the final reservoir states over $25$ streams drawn from all $2^{10}$ ten-bit streams, against $\kavg$ ($\nR{=}25$). Shared pool: the wiring of Snyder et al., where input and recurrent edges come from one pool. Separate draw: the input edges drawn on their own, as many on average as in the shared pool. Dedicated: the one-to-one mapping at input fan-out $\redund{=}1$, $3$ and $5$. Under the shared pool and the separate draw, few or no nodes receive the input at low $\kavg$; the dedicated mapping at $\redund{=}5$ raises KQ there. Lines: smoothing splines through the means over $25$ reservoirs at each $\kavg$; the numbers in the text are the means at the grid points.}
\label{fig:supp_snyder}
\end{figure}

We reproduce Snyder et al.'s kernel-quality curve and then change only the input wiring (Fig.~\ref{fig:supp_snyder}; \texttt{generate\_replication.py}). The setup follows theirs: $\nR{=}25$, one input node, $25$ streams drawn from all $2^{10}$ streams of $10$ bits, GR streams that share their last $\tau{=}3$ steps (Appendix~\ref{app:probe}), KQ and GR streams drawn independently, $25$ reservoirs per point, and no washout, as in the definition of Snyder et al. The replication reproduces the published S-curve on the shared pool with heterogeneous in-degrees, as in their Fig.~1a: KQ rises from $1.5$ of 25 at $\kavg{=}1$ to $24.9$ at $\kavg{=}8$ (homogeneous in-degrees: $2.2$ to $25.0$), and GR rises with KQ. Drawing the input edges separately from the recurrent ones, as many on average as in the shared pool, changes the curve little at low $\kavg$ (KQ $1.6$ against $1.5$ shared at $\kavg{=}1$, and $16.2$ against $14.3$ at $\kavg{=}3$), since at low $\kavg$ the input may still reach no node. The one-to-one mapping with $\redund{=}1$ connects the input to exactly one reservoir node at every $\kavg$, and KQ reaches only $16.6$ by $\kavg{=}8$; at $\redund{=}5$, KQ is $4.2$, $18.6$ and $25.0$ at $\kavg \in \{1,3,8\}$, above the shared pool at every $\kavg$.

\subsection{Statistical Checks}\label{app:stats}

Every number in Sect.~\ref{sec:results} and in Appendices~\ref{app:stats}--\ref{app:probe} is computed by \texttt{analyze\_design\_choices.py} in the released code (Sect.~\ref{sec:method_design}), which writes it to a CSV file; each check below names its file. Design accuracy is a median over a design's $20$ reservoirs, and the checks test whether the comparisons in the main text exceed the variation between reservoirs of the same design. Each reservoir belongs to one design only, so the checks resample reservoirs within each design or cluster standard errors by design, instead of fitting a random effect per design.

\emph{Precision of $\kbest$} (\texttt{k\_boot}). Each of $200$ draws resamples every design's $20$ reservoirs with replacement, uses the same resampled reservoirs for every task variant, and recomputes $\kbest$, the $\kavg$ at the peak of each mean curve. The designs are not resampled, since the grid contains every design. The $95\,\%$ interval of $\kbest$ is at most one grid step wide at every fan-out, readout and task except summed parity at fan-out $4$, where it spans two steps (Table~\ref{tab:supp_kboot}), so the fall of $\kbest$ with fan-out is not an effect of which reservoirs were drawn. The bars of Fig.~\ref{fig:kbest} show the range of $\kavg$ within $1\,\%$ of a curve's peak, not how precisely the peak is known: from fan-out $16$ up, the $\kavg$ within $1\,\%$ of a peak span $0.25$--$1$ (\texttt{k\_by\_drive}).

\emph{Each combination's $\kbest$} (\texttt{k\_design}). $\kbest$ comes from a curve averaged over the combinations of the other factors, and a single combination may be most accurate at another $\kavg$. Each combination's own $\kbest$, taken over the combinations whose accuracy varies by at least $2\,\%$ over $\kavg$, agrees: from fan-out $4$ up, its median over combinations equals $\kbest$ at $19$ of the $20$ points of Fig.~\ref{fig:kbest}, and lies $0.25$ below it, inside the interval, on final-state majority at fan-out $512$; no interquartile range is wider than $0.5$. On final-state parity at fan-out $4$ only $73$ of the $336$ curves ($84$ combinations, four task variants) vary by $2\,\%$, and every other combination keeps at least $154$. At fan-out $1$ the medians are $4$--$7$ (\texttt{k\_by\_drive\_low} gives the mean curves there).

\emph{Choosing on the selection run} (\texttt{selection\_drop}). The design chosen as best on the selection run is less accurate on the reported reservoirs, since part of its advantage on the selection run came from that run's reservoirs. Over the $8$ task variants of a readout, the mean drop is $3.4$--$10.5\,\%$ at fan-out $1$--$64$ and $0.4$--$1.7\,\%$ at $256$--$512$; the largest is $22.6\,\%$, at fan-out $1$. Within each task variant's best design, single-reservoir accuracy has a standard deviation of $1.4$--$19.5\,\%$, above $4\,\%$ in $14$ of the $16$ task variants (\texttt{draws}), against a sampling noise of $0.64\,\%$ between pattern sets (Sect.~\ref{sec:method_tasks}).

\emph{Task and delay} (\texttt{k\_task}, \texttt{k\_task\_summary}). Task pairs and delay pairs (Sect.~\ref{sec:results_kbest}), at the same readout, window and fan-out, are compared on the same reservoirs. Each of $200$ draws resamples every design's reservoirs once and recomputes $\kbest$ of both task variants; the two values of $\kbest$ differ beyond the variation between reservoirs when the $95\,\%$ interval of their difference excludes $0$. A pair is left out when either variant's own $\kbest$ has an interval wider than $1$. This rule leaves out three task pairs and three delay pairs, all with final-state parity at $d{=}10$ and fan-out $256$--$512$, where accuracy stays at chance; the intervals of their differences are $3$--$4.25$ wide, and every other interval of a difference is at most $0.75$ wide.

\emph{Other factors and $\kbest$} (\texttt{k\_shift}). The shifts of $\kbest$ with injection, ticks, initialization and self-loops are differences between the $\kavg$ at the peaks of mean curves, one curve per level, and have no interval. At $\selfloop{=}1$ a node without in-edges has no source for its self-loop to replace, so it first receives an edge moved from a node with two or more in-edges. The rise of $\kbest$ does not depend on these moved edges: already at $\selfloop{=}0.5$, where no edge is moved, $\kbest$ rises by $0$--$0.5$, and the mean number of edges from other nodes at $\kbest$, $\kbest - \selfloop$, falls or stays equal in all $20$ combinations. The rise stays below one edge per node and is not proportional to $\selfloop$: at $\selfloop{=}1$, $\kbest$ rises by $0.75$, one grid step less than one edge per node, in $7$ of the $20$ combinations, and from $\selfloop{=}0.5$ to $1$ it rises by $0.5$, as many edges per node as the added self-loops replace, in $8$.

\emph{Homogeneous wiring} (\texttt{wiring\_k}, \texttt{wiring\_best}, \texttt{wiring\_best\_summary}). The second grid, in the homogeneous mode, has no selection run, so each wiring mode's best design over injection and $\kavg \in \{1, \dots, 8\}$ is chosen on $10$ of a design's reservoirs and evaluated on the other $10$, then the other way round, and the two accuracies are averaged. The homogeneous and the heterogeneous mode choose among the same $16$ designs; the heterogeneous reservoirs are those of the main grid. Each of $200$ draws resamples each half's reservoirs within each design, so choosing and evaluating stay disjoint. The mean difference, homogeneous minus heterogeneous, over the $16$ task variants is $+1.1$, $+3.7$, $+4.1$, $+2.7$, $+2.0$ and $+0.4\,\%$ at fan-out $1$ to $512$; the $95\,\%$ interval of this mean lies above $0$ at fan-out $64$ ($0.3$ to $4.6\,\%$) and $256$ ($1.2$ to $2.9\,\%$), and is $7$--$10\,\%$ wide at fan-out $1$--$16$. Per task variant and fan-out, $24$ of the $96$ intervals lie above $0$ and none below.

\emph{Variance shares} (\texttt{variance}, \texttt{draw\_share}). Over the whole grid, input fan-out and $\kavg$ account for $42$--$86\,\%$ of the variance in design accuracy, depending on the task variant. A large share means that a factor separates good designs from bad ones, not that it changes accuracy among good designs, which is why the other factors are compared at the operating point. Of the other factors, injection accounts for at most $21\,\%$, and ticks, initialization and self-loops for at most $7\,\%$ each. The differences between reservoirs of the same design account for more than half of the variance of single-reservoir accuracy on three task variants of final-state parity ($w{=}5$ at $d{=}5$, and both windows at $d{=}10$).

\emph{Other factors} (\texttt{twins\_summary}, \texttt{self\_loops\_summary}, \texttt{beta\_op}). At a task variant's operating point, ticks, initialization and self-loops form a full $2 \times 3 \times 7$ factorial. Per readout and task we fit a Beta regression (logit link) of single-reservoir accuracy on the task variant and the three factors, as main effects relative to one tick, random initialization and no self-loops, with standard errors clustered by design ($3{,}360$ reservoirs per fit). The fit for ticks and initialization uses the operating point's $\kavg$, with self-loops as a control; the fit for self-loops places each combination at its own $\kbest$, with ticks and initialization as controls (Sect.~\ref{sec:method_design}). Table~\ref{tab:supp_beta} gives each factor's average marginal effect, the mean change in predicted accuracy when only that factor changes. The model estimates mean effects, whereas Table~\ref{tab:choices} reports medians of paired differences, so the sizes differ, but the conclusions agree: a second tick, and self-loops on $20\,\%$ or more of the nodes, lower accuracy on every readout and task; the $95\,\%$ intervals of self-loops on $10\,\%$ of the nodes ($s{=}0.1$) and of either initialization include $0$ on every readout and task.

\begin{table}[tb]
\centering
\small
\caption{Average marginal effect of each factor at the operating point, in $\%$, from a Beta regression per readout and task (logit link; baseline one tick, random initialization, no self-loops; standard errors clustered by design). Ticks and initialization: at the operating point's $\kavg$, with self-loops as a control. Self-loops: each combination at its own $\kbest$, with ticks and initialization as controls. Brackets: $95\,\%$ interval from $2{,}000$ coefficient draws.}
\label{tab:supp_beta}
\begin{tabular}{lcccc}
\toprule
 & final-state & final-state & summed & summed \\
 & majority & parity & majority & parity \\
\midrule
ticks $\nu{=}2$       & $-13.0$ & $-5.8$ & $-8.0$ & $-2.8$ \\
                      & {\scriptsize$[-15.1, -10.8]$} & {\scriptsize$[-6.9, -4.8]$} & {\scriptsize$[-9.7, -6.3]$} & {\scriptsize$[-3.5, -2.2]$} \\[3pt]
self-loops $s{=}0.1$  & $-5.4$ & $-2.3$ & $-2.2$ & $-0.7$ \\
                      & {\scriptsize$[-10.6, +0.2]$} & {\scriptsize$[-4.7, +0.1]$} & {\scriptsize$[-5.6, +1.6]$} & {\scriptsize$[-2.2, +0.8]$} \\[3pt]
self-loops $s{=}0.2$  & $-9.3$ & $-3.5$ & $-3.8$ & $-1.8$ \\
                      & {\scriptsize$[-13.9, -4.1]$} & {\scriptsize$[-6.0, -0.9]$} & {\scriptsize$[-6.7, -0.4]$} & {\scriptsize$[-3.2, -0.4]$} \\[3pt]
self-loops $s{=}0.3$  & $-11.2$ & $-4.4$ & $-4.7$ & $-1.7$ \\
                      & {\scriptsize$[-15.6, -6.1]$} & {\scriptsize$[-6.7, -2.1]$} & {\scriptsize$[-7.8, -1.2]$} & {\scriptsize$[-3.2, -0.3]$} \\[3pt]
self-loops $s{=}0.4$  & $-12.7$ & $-5.0$ & $-6.4$ & $-2.4$ \\
                      & {\scriptsize$[-17.3, -7.5]$} & {\scriptsize$[-7.3, -2.7]$} & {\scriptsize$[-9.8, -2.8]$} & {\scriptsize$[-3.8, -1.0]$} \\[3pt]
self-loops $s{=}0.5$  & $-13.9$ & $-5.4$ & $-6.9$ & $-2.7$ \\
                      & {\scriptsize$[-18.6, -8.7]$} & {\scriptsize$[-7.6, -3.1]$} & {\scriptsize$[-9.9, -3.3]$} & {\scriptsize$[-4.1, -1.2]$} \\[3pt]
self-loops $s{=}1.0$  & $-15.5$ & $-5.6$ & $-8.2$ & $-3.3$ \\
                      & {\scriptsize$[-20.6, -9.9]$} & {\scriptsize$[-8.1, -3.2]$} & {\scriptsize$[-12.1, -3.8]$} & {\scriptsize$[-4.8, -1.8]$} \\[3pt]
init zeros            & $-1.2$ & $-0.2$ & $-0.4$ & $-0.3$ \\
                      & {\scriptsize$[-4.2, +1.9]$} & {\scriptsize$[-1.6, +1.2]$} & {\scriptsize$[-2.7, +1.8]$} & {\scriptsize$[-1.1, +0.5]$} \\[3pt]
init random-warmup    & $-0.7$ & $0.0$ & $-1.0$ & $-0.4$ \\
                      & {\scriptsize$[-4.0, +2.4]$} & {\scriptsize$[-1.2, +1.3]$} & {\scriptsize$[-3.1, +1.2]$} & {\scriptsize$[-1.2, +0.4]$} \\
\bottomrule
\end{tabular}
\end{table}

\begin{table}[tb]
\centering
\small
\caption{The optimal connectivity $\kbest$ per input fan-out (the $\kavg$ at the peak of the mean curve, as in Fig.~\ref{fig:kbest}). Brackets: $95\,\%$ interval over $200$ bootstrap draws of each design's reservoirs.}
\label{tab:supp_kboot}
\begin{tabular}{lcccc}
\toprule
fan-out & final-state & final-state & summed & summed \\
 & majority & parity & majority & parity \\
\midrule
$4$ & $2.5$\,{\scriptsize$[2.5, 2.5]$} & $3$\,{\scriptsize$[2.75, 3]$} & $2.5$\,{\scriptsize$[2.5, 2.5]$} & $2.75$\,{\scriptsize$[2.5, 3]$} \\
$16$ & $2.25$\,{\scriptsize$[2.25, 2.25]$} & $2.25$\,{\scriptsize$[2.25, 2.25]$} & $2.25$\,{\scriptsize$[2.25, 2.25]$} & $2.25$\,{\scriptsize$[2.25, 2.25]$} \\
$64$ & $2.25$\,{\scriptsize$[2, 2.25]$} & $2.25$\,{\scriptsize$[2, 2.25]$} & $2$\,{\scriptsize$[2, 2]$} & $2$\,{\scriptsize$[2, 2]$} \\
$256$ & $2$\,{\scriptsize$[2, 2]$} & $2$\,{\scriptsize$[2, 2]$} & $1.75$\,{\scriptsize$[1.75, 2]$} & $1.75$\,{\scriptsize$[1.75, 1.75]$} \\
$512$ & $2$\,{\scriptsize$[1.75, 2]$} & $2$\,{\scriptsize$[2, 2]$} & $1.75$\,{\scriptsize$[1.75, 1.75]$} & $1.75$\,{\scriptsize$[1.75, 1.75]$} \\
\bottomrule
\end{tabular}
\end{table}

\subsection{The Probe}\label{app:probe}

\emph{Probe settings} (\texttt{settings\_summary}). $\mathcal{U}_{\tau}$ copies $\tau$ pattern bits from one reference stream, at the same positions in every stream. The \emph{placement} sets the positions: the first $\tau$ pattern bits (\emph{first}), the last $\tau$ (\emph{last}), or $\tau$ pattern bits drawn at random (\emph{random}). In \emph{steal} mode the copied bits replace pattern bits; in \emph{augment} mode every stream has $\tau$ more pattern bits, so all $15$ original pattern bits keep their own values, and $\mathcal{S} = \{11, \dots, T{+}\tau\}$. The $m$ streams of $\mathcal{U}_{\tau}$ are drawn at random after duplicate streams are removed, and $\mathcal{U}$ holds the same $m$ streams before the copy. A \emph{probe setting} is one combination of placement, $\tau \in \{3,5\}$ and mode, $12$ in all; larger $\tau$ is not used, since $\tau \ge 7$ leaves fewer than $512$ distinct streams for the GR rank. The \emph{probe} means KQ, GR and $\Delta$ under one probe setting. Every reservoir is measured under all $12$. Last + steal with $\tau{=}5$ was chosen on the selection run. On the reported reservoirs, only under last + steal, with $\tau{=}3$ or $5$, does $\Delta$ correlate with single-reservoir accuracy at the operating point, within each self-loop level, on summed parity ($\rho = 0.37$--$0.39$, averaged over its four task variants, against at most $0.11$ for the other ten settings). Why only last + steal correlates on summed parity is open.

\emph{Analysis.} KQ and GR are compared with accuracy only within groups of equal $\kavg$ and equal self-loops, never across them, since $\kavg$ changes KQ, GR and accuracy at once, and a self-loop, like a change of $\kavg$, changes the number of edges between distinct nodes. At a fixed $\kavg$, $\Delta$ falls with self-loops, as accuracy does, so across all $42$ designs of an operating point a correlation of $\Delta$ with accuracy would mostly measure the effect of self-loops.

\emph{Locating $\kbest$} (\texttt{peaks}, \texttt{peaks\_s}). KQ and GR rise up to $\kavg{=}8$ at every fan-out (the S-curve, Sect.~\ref{sec:background}); only $\Delta$ peaks inside the $\kavg$ range. For each readout and task, the $\kavg$ at $\Delta$'s peak lies within $0.5$ of each task variant's own $\kbest$ in $85$--$100\,\%$ of the combinations of fan-out and task variant, and a fixed $\kavg{=}2$ in $65$--$100\,\%$. The only cases where neither the $\kavg$ at $\Delta$'s peak nor $\kavg{=}2$ is within $0.5$ are final-state parity at $d{=}10$ at fan-out $256$ and $512$, whose mean curve stays at chance. At every self-loop level the $\kavg$ at $\Delta$'s peak lies within $0.5$ of $\kbest$ in $95$--$100\,\%$ of the combinations of readout, task and fan-out, and $\kavg{=}2$ in $85$--$100\,\%$ below $s{=}1$ and in $50\,\%$ at $s{=}1$.

\emph{Ranking designs} (\texttt{design\_rho}, \texttt{design\_rho\_k\_step}). Within each self-loop level at the operating point, design accuracy differs mainly with ticks, since the initialization changes it by at most $0.5\,\%$. Final-state parity at $d{=}10$ is left out here: its design accuracies are $51$--$59\,\%$, close to chance. At the chosen operating points, a design's mean $\Delta$ correlates positively with design accuracy (permutation $p < 0.05$) on $9$ of the other $14$ task variants (row 1 of Table~\ref{tab:kqgr}): all four of summed majority, three of summed parity and final-state majority at $d{=}5$. On final-state majority at $d{=}10$ the correlation is negative ($\rho = -0.48$ and $-0.38$, $p = 0.003$ and $0.03$); on final-state parity at $d{=}5$ and on summed parity at $w{=}5$, $d{=}10$ there is none ($|\rho| \le 0.21$, $p \ge 0.23$). These correlations depend on the operating point's $\kavg$: one grid step lower, $\rho$ is positive on final-state majority at $d{=}10$ ($0.37$ and $0.30$) and on both final-state parity variants at $d{=}5$, and one step higher it is negative on those parity variants. On final-state majority at $d{=}10$, the negative correlation comes from the second tick, which raises $\Delta$ while it lowers accuracy (Fig.~\ref{fig:supp_delta}): on final-state majority at $w{=}5$, $d{=}10$ without self-loops, a second tick raises $\Delta$ from $21$ to $30$ and lowers design accuracy from $67\,\%$ to $55\,\%$ (means over the three initializations). A possible reading, not tested here: two updates per input bit make the states of different streams differ more, which raises $\Delta$, and keep fewer past input bits in the state, which $\Delta$ does not measure. KQ and GR correlate positively with design accuracy on all $14$ task variants.

\emph{Ranking reservoirs} (\texttt{seeds}, \texttt{seeds\_recommended}). Within a design, $\Delta$ correlates with accuracy, but the reservoir with the highest $\Delta$ is often not the most accurate: the $\rho$ between $\Delta$ and accuracy across the reservoirs of one design is positive on $15$ of the $16$ task variants (row 2 of Table~\ref{tab:kqgr}), but on the operating point's three designs with one tick and no self-loops, the reservoir with the highest $\Delta$ is more accurate than the median one on $10$ task variants, up to $4.6\,\%$ less accurate on $5$, all on the final-state readout, and equal on one (row 3).

\begin{table}[tb]
\centering
\small
\setlength{\tabcolsep}{4pt}
\caption{$\Delta$ against accuracy, under last + steal with $\tau{=}5$; each entry is a range over the column's four task variants. Row 1: the Spearman $\rho$ between a design's mean $\Delta$ over its reservoirs and design accuracy, across the operating point's designs with the same fraction of self-loops (six: ticks $\times$ initialization), averaged over the seven self-loop levels. Row 2: $\rho$ across the reservoirs of one design, the mean over the operating point's $42$ designs. Rows 3--4: accuracy minus that of the design's median reservoir, in $\%$, for the reservoir with the highest $\Delta$ (row 3) and the most accurate reservoir (row 4), the mean over the operating point's three designs with one tick and no self-loops (one per initialization); row 3 averages tied reservoirs, since $\Delta{=}0$ for $30\,\%$ of final-state reservoirs and $74\,\%$ of summed ones.}
\label{tab:kqgr}
\begin{tabular}{lcccc}
\toprule
 & final-state & final-state & summed & summed \\
 & majority & parity & majority & parity \\
\midrule
across designs           & $-0.48$--$0.79$ & $-0.09$--$0.79$ & $0.72$--$0.80$ & $-0.21$--$0.81$ \\
within designs           & $0.26$--$0.56$ & $0.16$--$0.49$ & $0.43$--$0.48$ & $-0.01$--$0.21$ \\
\midrule
highest $\Delta$         & $-3.6$--$9.8$ & $-4.6$--$6.2$ & $2.6$--$7.2$ & $0.0$--$8.5$ \\
most accurate            & $9.8$--$25.6$ & $14.5$--$23.2$ & $13.2$--$18.0$ & $2.2$--$33.1$ \\
\bottomrule
\end{tabular}
\end{table}

\clearpage
\subsection{Supplementary Scatters}\label{app:scatters}

\begin{figure}[h]
\centering
\begin{minipage}[t]{0.49\linewidth}
\centering
\dashfig{/furu/design/}{raw_acc_by_k_density}{\includegraphics[width=\linewidth]{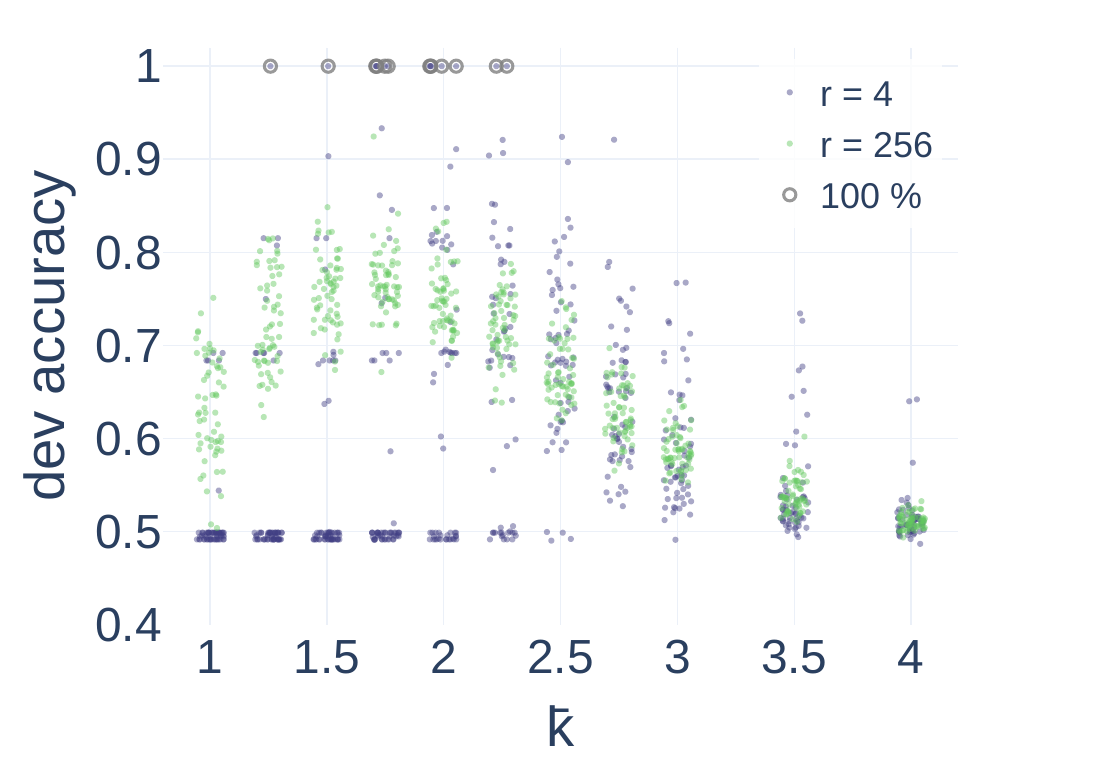}}\\[2pt]
(a)
\end{minipage}\hfill
\begin{minipage}[t]{0.49\linewidth}
\centering
\dashfig{/furu/design/}{raw_acc_by_k_parity}{\includegraphics[width=\linewidth]{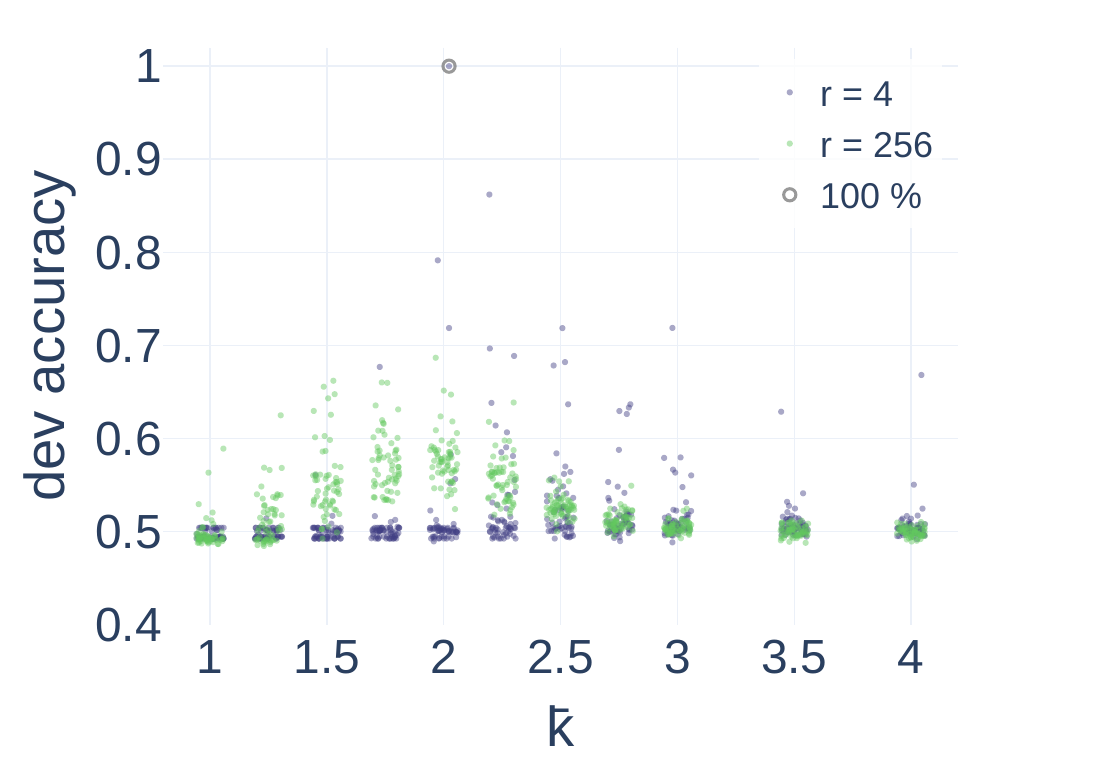}}\\[2pt]
(b)
\end{minipage}
\caption{Single-reservoir accuracy against $\kavg$ for each input fan-out, final-state readout, one point per reservoir. (a)~Majority: at $\redund{=}256$ accuracy is highest at a lower $\kavg$ than at $\redund{=}4$; the vertical spread at one $\kavg$ comes from the other factors and from the variation between reservoirs of the same design. (b)~Parity: the same picture, with more reservoirs near $50\,\%$, many of them at $\redund{=}4$.}\label{fig:supp_k}
\end{figure}

\begin{figure}[h]
\centering
\begin{minipage}[t]{0.49\linewidth}
\centering
\dashfig{/furu/analysis/}{design_op_delta_by_acc_density_w5_d5}{\includegraphics[width=\linewidth]{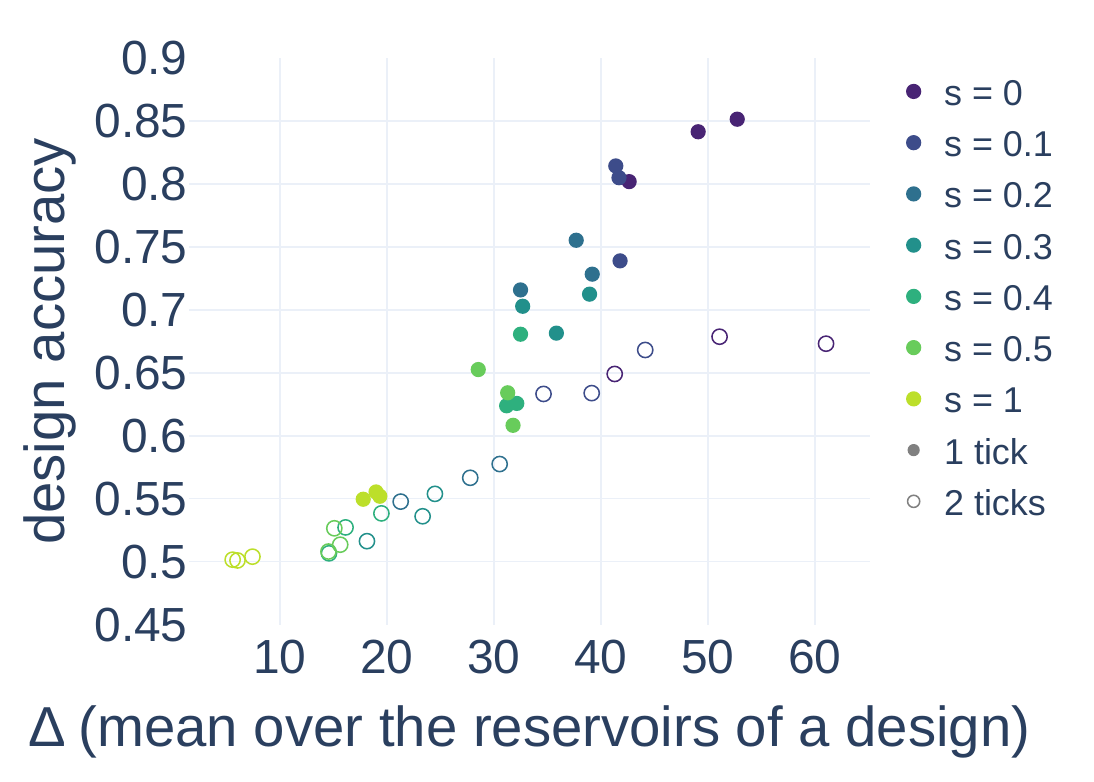}}\\[2pt]
(a)
\end{minipage}\hfill
\begin{minipage}[t]{0.49\linewidth}
\centering
\dashfig{/furu/analysis/}{design_op_delta_by_acc_density_w5_d10}{\includegraphics[width=\linewidth]{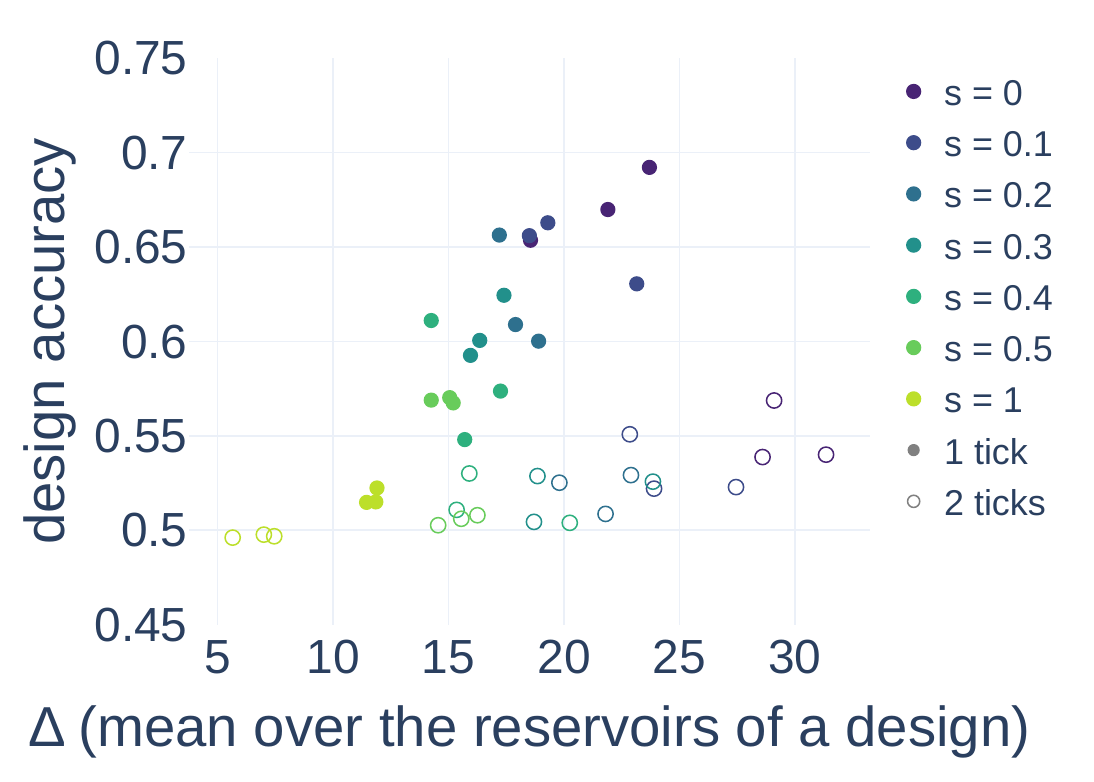}}\\[2pt]
(b)
\end{minipage}
\caption{$\Delta$ against design accuracy at the operating point, final-state majority at $w{=}5$, one point per design; colour gives the fraction $s$ of nodes with a self-loop, filled markers one tick and open markers two. (a)~Delay $5$: $\Delta$ and accuracy fall together as $s$ rises, and within each $s$ the one-tick designs are the more accurate; within an $s$ level $\rho = 0.74$, the mean over the levels. (b)~Delay $10$: below $s{=}1$ the second tick raises the mean $\Delta$ of a level and lowers its accuracy, so within an $s$ level $\Delta$ ranks the designs in reverse ($\rho = -0.38$, the mean over the levels).}\label{fig:supp_delta}
\end{figure}

\begin{figure}[h]
\centering
\begin{minipage}[t]{0.49\linewidth}
\centering
\dashfig{/furu/design/}{raw_test_by_dev}{\includegraphics[width=\linewidth]{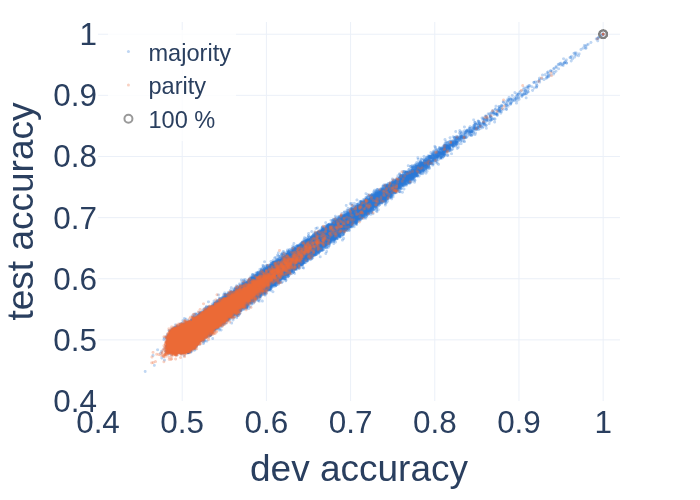}}
\end{minipage}
\caption{Test against development accuracy, final-state readout at $w{=}5$, $d{=}5$, one point per reservoir; reservoirs at $100\,\%$ on the development set stay at $100\,\%$ or just below on test.}\label{fig:supp_devtest}
\end{figure}

Figures~\ref{fig:supp_k}--\ref{fig:supp_devtest} show the single designs and reservoirs behind the summaries in the main text. Each caption states whether a point is a reservoir or a design; rings mark reservoirs at $100\,\%$ accuracy.

